\documentclass{article}

\usepackage{microtype}
\usepackage{graphicx}
\usepackage{subfigure}
\usepackage{booktabs} 

\usepackage{hyperref}

\usepackage[accepted]{icml2023}

\usepackage{amsmath}
\usepackage{amssymb}
\usepackage{mathtools}
\usepackage{amsthm}

\usepackage[capitalize,noabbrev]{cleveref}

\usepackage{bm}
\usepackage{tabularx}
\usepackage{xurl}

\theoremstyle{plain}

\theoremstyle{definition}

\theoremstyle{remark}

\usepackage[textsize=small]{todonotes}

\begin{document}

\twocolumn[
\icmltitle{A Fair Loss Function for Network Pruning}



\icmlsetsymbol{equal}{*}

\begin{icmlauthorlist}
\icmlauthor{Robbie Meyer}{uwaterloo}
\icmlauthor{Alexander Wong}{uwaterloo}
\end{icmlauthorlist}

\icmlaffiliation{uwaterloo}{Department of Systems Design Engineering, University of Waterloo, Waterloo, Canada}

\icmlcorrespondingauthor{Robbie Meyer}{robbie.meyer@uwaterloo.ca}

\icmlkeywords{Pruning, Fairness, Compression, Image Classification, Machine Learning}

\vskip 0.3in
]



\printAffiliationsAndNotice{}  

\begin{abstract}
  Model pruning can enable the deployment of neural networks in environments with resource constraints. While pruning may have a small effect on the overall performance of the model, it can exacerbate existing biases into the model such that subsets of samples see significantly degraded performance. In this paper, we introduce the performance weighted loss function, a simple modified cross-entropy loss function that can be used to limit the introduction of biases during pruning. Experiments using the CelebA, Fitzpatrick17k and CIFAR-10 datasets demonstrate that the proposed method is a simple and effective tool that can enable existing pruning methods to be used in fairness sensitive contexts. Code used to produce all experiments contained in this paper can be found at \url{https://github.com/robbiemeyer/pw_loss_pruning}.
\end{abstract}

\section{Introduction}\label{sec:intro}

Deep learning models are large, requiring millions of operations to make an inference \cite{canziani2017}. Deploying large neural networks to environments with limited computational resources, such as mobile and embedded devices, may be infeasible.

Pruning is a simple and common method for reducing the size of a neural network \cite{blalock}. It involves identifying parameters that do not significantly affect the model's output and removing them from the network. Pruning enables the deployment of performant neural networks in resource constrained environments \cite{wu, vemparala}. However, recent research has shown that while overall accuracy of the model may be maintained while the model is compressed, pruning can exacerbate existing model biases, disproportionately affecting disadvantaged groups \cite{hooker}. Pruning methods that are designed to preserve overall model performance may not prioritize the preservation of parameters that are only important for a small subset of samples.

This effect has significant implications for the implementation of pruning in real-world situations. Biases have been observed in artificial intelligence systems such as those used to classify chest X-ray images \cite{seyyed}, recognize faces \cite{snow} and screen resumes \cite{dastin}. Biases in models can increase the risk of unfair outcomes, preventing the implementation of the model. If pruning exacerbated a model's biases, it could increase the risk of unfair outcomes or limit the deployment of the pruned model. It is therefore important to prune in a manner that does not aggravate a model's biases.

In this paper, we propose the performance weighted loss function as a simple method for boosting the fairness of data-driven methods for pruning convolutional filters in convolutional neural network image classifiers. The goal of our method is to enable the pruning of a significant number of model parameters without significantly exacerbating existing biases. The loss function consists of two small tweaks to the standard cross-entropy loss function to prioritize the model's performance for poorly-classified samples over well-classified samples. These tweaks can be used to extend existing data-driven pruning methods without requiring explicit attribute information.

We demonstrate the effectiveness of our approach by pruning classifiers using two different pruning approaches for the CelebA \cite{celeba}, Fitzpatrick17k \cite{fitzpatrick} and CIFAR-10 \cite{cifar10} datasets. Our results show that the performance weighted loss function can enable existing pruning methods to prune neural networks without significantly increasing model bias. 

\section{Related Work}\label{sec:relworks}

Many different pruning approaches have been proposed to reduce the size of CNNs while minimally impacting model accuracy. Pruning methods typically involve assigning a score to each parameter or group of parameters, removing parameters based on these scores and retraining the newly pruned network to recover lost accuracy \cite{blalock}. 

The procedure by which parameters are identified to be pruned is the primary differentiator between pruning methods. There are a wide variety of scoring approaches used to identify parameters that are unimportant or redundant and can be removed from the network. Many approaches use parameter magnitudes to identify parameters to prune \cite{li, han}. Other approaches use gradient information \cite{liu}, Taylor estimates of parameter importance \cite{molchanov2017, molchanov, ide2020} and statistical properties of future layers \cite{luo}. Some approaches involve learning the scores via parameters that control the flow of information through the network \cite{castells, you}.

However, almost all novel pruning approaches focus on the overall accuracy of the model after pruning. There are few papers that aim to improve or analyze the fairness of a pruned model. Hooker et al. \cite{hooker} propose auditing samples affected by model compression, called Compression Identified Exemplars, as an approach for identifying and managing the negative effects of model compression. \citet{paganini2020} demonstrates how class fairness can be affected by pruning approaches that only consider overall accuracy. \citet{fairprune} propose Fairprune, a method for improving model bias using pruning. Instead of seeking to compress a model, Fairprune prunes parameters using a saliency metric to increase model fairness. \citet{xu2022} propose the use of knowledge distillation and pruning to reduce bias in natural language models. \citet{joseph2020} propose a multi-part loss function intended to improve the alignment between predictions between the original and pruned model. They demonstrate that their method can have beneficial effects for fairness between classes. \citet{marcinkevivcs2022} propose a debiasing procedure that involves pruning parameters using a gradient based influence measure.

While not a pruning method, the work of \citet{mahabadi2019} is also relevant as their ``Debiased Focal Loss'' resembles the weighting scheme of the loss function proposed in this paper. Instead of using their loss function for model compression, they aim to debias a model using the output of a trained bias-only model.

\section{Method}\label{sec:method}

\subsection{Motivation}

In the unfair pruning situation described by \citet{hooker}, model performance was more significantly impacted for certain sample subgroups. The highly impacted subgroups were characterized by poor representation in the training data and worse subgroup performance by the original model when compared to unimpacted groups. The performance decrement induced by the pruning process disproportionately impacts subgroups which are underrepresented and poorly classified.

To rectify this inequality, we can design a pruning process that prioritizes maintaining the performance of samples from the impacted subgroups. However, we do not need to develop a new pruning method from scratch to achieve this objective. Many existing pruning methods use data to identify which model parameters should be removed. Some methods use parameters learned via a loss minimization process whereas others values derived from gradients calculated with respect to a loss function. By modifying the loss function to prioritize samples from impacted subgroups, we can boost the fairness of existing pruning methods.

\subsection{The Performance Weighted Loss Function}

We make two different modifications to the standard cross-entropy loss function to transform it into the performance weighted loss function (PW loss). We first apply sample weighting to ensure that samples from impacted groups have a larger contribution to the loss function. We then transform the sample labels to ensure that we are not reinforcing undesirable model behaviours.

As the attribute information required to identify impacted subgroups is not always readily accessible, our weighting scheme does not depend on any external information. We instead use the output of the original model to determine each sample weight. We assign larger weights to samples for which the original model was not able to confidently classify. The form of the scheme resembles the focal loss \cite{lin2017}. However, as the samples are weighted using the outputs of the original model the weights do not depend on the current output of the model and will not change during training. The weight assigned to the $i$th data sample, $w_i$, is given by the following equation: 
\begin{equation}\label{eq:weights}
  w_i = \theta + (1 - \hat{y}_i)^\gamma 
\end{equation}
where $\hat{y}_i$ is the predicted probability given by the original model for the sample's true class, $\theta \in [0,1]$ is the minimum weight value and $\gamma \ge 0$ controls the shape of the relation between $\hat{y}_i$ and $w_i$.

We also emphasize the model performance through the use of corrected soft-labels in the cross-entropy function. Rather than using the true labels of each sample, we use the output of the original model for the loss function in the pruning process. Without this change, the preservation of an originally poorly classified sample's prediction probability would result in a greater loss value than the preservation of an originally well classified sample's prediction probability. The use of true labels implicitly prioritizes the preservation of model performance for samples that have predictions closer to their true labels. Using the model output as soft-labels alleviates this implicit prioritization.

However, as we are assigning higher weights to samples that are originally classified by the original model while also using the original model's output as our labels, we are consequently assigning the highest weights to incorrect labels. To avoid emphasizing incorrect behaviours we correct the soft-labels. The corrected soft-label, $\bm{\hat{y}}^*_i$ is defined as: 
\begin{equation}
  \bm{\hat{y}}^*_i = \begin{cases}
    \bm{\hat{y}}_i \quad \textrm{if} \quad \hat{C}_{i} = C_{i} \\
    \bm{y}_i \quad \textrm{otherwise} \\
  \end{cases}
\end{equation}
where $\bm{\hat{y}}_i$ contains the prediction probabilities derived from the model output for the $i$th sample, $\bm{y}_i$ is the true label vector of the $i$th sample, $\hat{C}_{i}$ is predicted class of the $i$th sample and $C_i$ is the true class of the $i$th sample. The corrected soft-label takes on the value of the model's prediction probabilities when the prediction is correct and the true label when the prediction is incorrect.

By the application of the performance weighted scheme and corrected soft-labels onto the standard cross-entropy function, the performance weighted loss function, $\mathcal{L}_{PW}$, is defined by:
\begin{equation}
  \mathcal{L}_{PW} = \sum_{i=1}^N w_i l_{CE}(\bm{\hat{y}}^*_i, \bm{\hat{y}}'_i)
\end{equation}
where $\bm{\hat{y}}'_i$ contains the prediction probabilities derived from the model output for the $i$th sample after pruning, $l_{CE}(\bm{\hat{y}}^*_i, \bm{\hat{y}}'_i)$ is the cross-entropy between the corrected soft-label and the prediction probabilities of the pruned model for the $i$th sample, and $N$ is the batch size.

By using this loss function with existing data-driven pruning methods, we can reduce the bias exaggerating effect of pruning by emphasizing samples that are more likely to be negatively affected by pruning.

\section{Experiments}\label{sec:exp}

\subsection{Experimental Set-up}

We applied the PW loss to two different pruning methods. The first method is AutoBot \cite{castells}, an accuracy preserving pruning method that uses trainable bottleneck parameters that limits the flow of information through the model. The second method uses an importance metric derived from the Taylor expansion of the loss function \cite{molchanov2017}. In both of our implementations, we pruned whole convolutional filters rather than individual neurons. As pruned filters can be fully removed from the model, rather than being set to zero, filter pruning is a simple method for directly reducing the FLOPS of a model.

In the AutoBot method, the bottlenecks are optimized by minimizing a loss function that includes the cross-entropy between the original and pruned model outputs, as well as terms that encourage the bottlenecks to limit information moving through the model, achieving a target number of FLOPS \cite{castells}. We applied the performance weighted loss function to the method by replacing the cross-entropy term in the loss function with the performance weighted loss function. Additionally, we also used the performance weighted loss function when retraining the model after pruning.

The importance metric of the Taylor expansion method is formed using the gradient of the loss function with respect to each feature map and the value of each feature map \cite{molchanov2017}. This method alternates between training the network and pruning a filter. In our implementation, a filter is pruned every five iterations. We applied the performance weighted loss function by replacing the loss functions used in the gradient calculation and model training with the performance weighted loss function. Once again, we also used the performance weighted loss function when retraining the model after pruning.

We also evaluated a random pruning method in which filters are selected and pruned from the network until only the desired number of FLOPS remain. We use the random pruning method as a reference.

We implemented the methods using the \textit{PyTorch} library \cite{pytorch}. The methods were implemented as three step pipelines in which the model is first pseudo-pruned by setting parameters to zero, fully pruned using the \textit{Torch-Pruning} library \cite{torchpruning} and retrained. Pseudo-pruning allows for fast pruning during the pruning process while the full pruning step removes the unused parameters, reducing the number of operations required for prediction. Due to dependencies between parameters introduced by structures such as residual layers, the achieved theoretical speedup often slightly differs from the target theoretical speedup. All hyperparameters for the pruning methods were selected using a hold-out validation set. Hyperparameters for the pruning methods were selected without the PW loss applied and were used for both unmodified and PW loss method variants. We repeated each experiment three times. All figures displaying model performance after pruning are displaying the average of all trials.

\begin{figure*}[t]
  \centering
  \includegraphics[width=\textwidth]{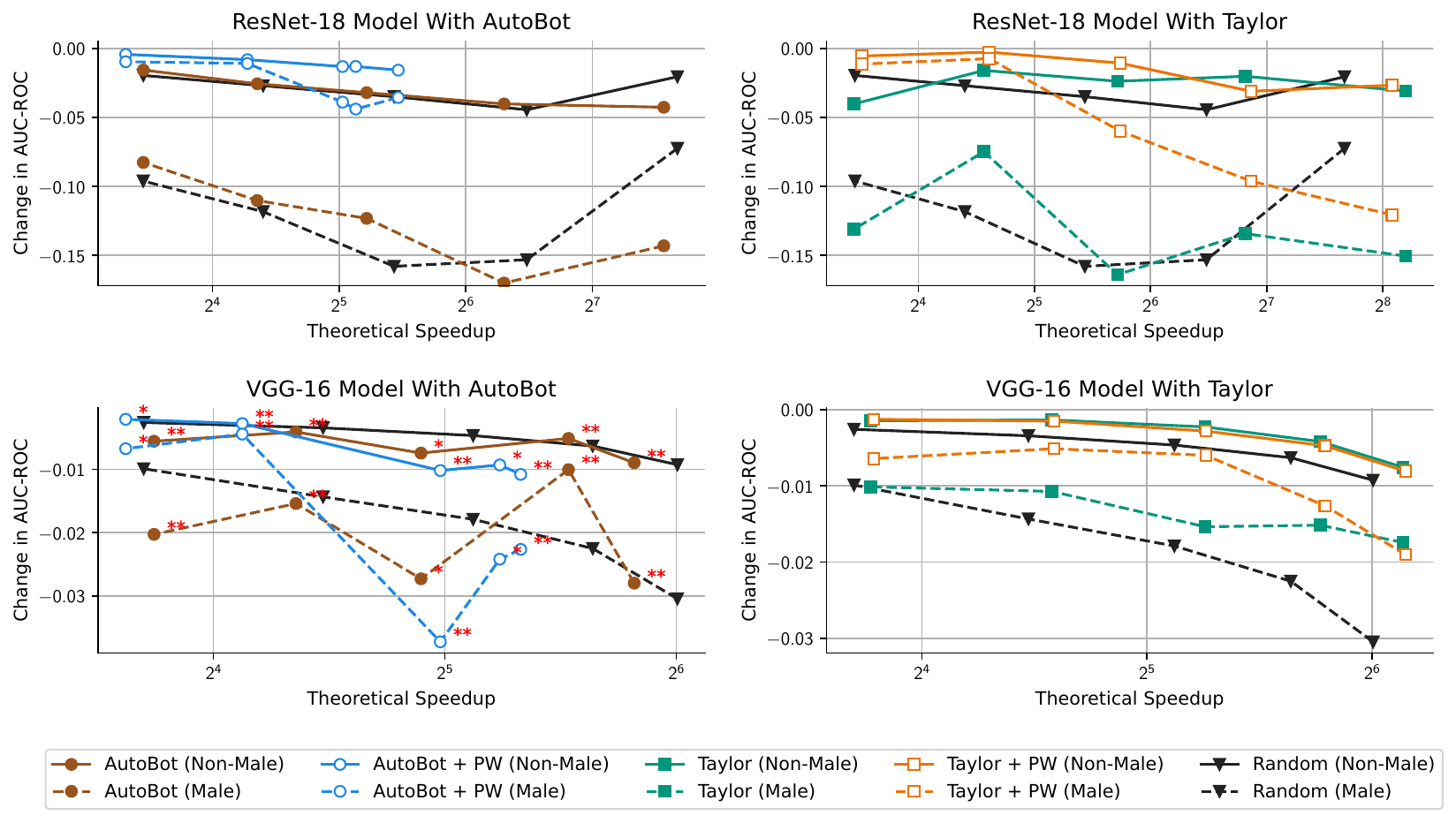}
  \vskip -0.02in
  \caption{Mean pruning performance with ResNet-18 and VGG-16 models with CelebA dataset. Red stars indicate degenerate trials.}
  \label{fig:celeba}
\end{figure*}

\subsubsection{Metrics}

Our primary concern is the degradation of a model's behaviour towards different subgroups due to pruning. We therefore evaluated the models by comparing the change in the areas under the receiver operator curves (ROC-AUC) for various subgroups for five different degrees of pruning. As it is a threshold agnostic performance metric, the ROC-AUC is a good measure of the model's understanding and separability for a subgroup \cite{borkan2019}. For non-binary classification we used the one-vs-one ROC-AUC.

We measured the degree to which a model is pruned using the theoretical speedup, defined as the FLOPS of the original model divided by the FLOPS of the pruned model.

\subsection{Evaluating Fairness and Performance}\label{sec:eval}

All methods were tested with and without the PW loss on three different classification tasks.

Our first task was the celebrity face classification task using the CelebA dataset \cite{celeba} as outlined by \citet{hooker}, in which a model is trained to identify faces as blonde or non-blonde. The CelebA dataset contains over 200 000 images of celebrity faces with various annotations. While blonde non-male samples make up 14.05\% of the training data, blond male samples make up only 0.85\% of the training data. We used the provided data splits with 80\% of the available data being used for training with the remaining data split evenly for validation and testing.

Our second task was the skin lesion classification task using the Fitzpatrick17k dataset \cite{fitzpatrick}. The Fitzpatrick17k dataset consists of 16 577 images of skin conditions. We trained our models to classify the samples as non-neoplastic, benign or malignant. Due to missing and invalid images we were only able to use 16 526 images. Each sample in the dataset is assigned a Fitzpatrick score that categorizes the skin tone of the sample. We trained our models on only samples with light skin tone scores of 1 or 2, and evaluated the model on medium skin tone scores of 3 or 4 as well as dark skin tone scores of 5 or 6. We used a random 25\% of the medium and dark skin tones as a validation set with the remainder used as a test set. As the skin tones the model is trained on and evaluated on differ, this is an out-of-distribution task.

Our final task was the CIFAR-10 classification task \cite{cifar10}. The CIFAR-10 dataset consists of 50 000 training images and 10 000 testing images evenly distributed between 10 classes. Unlike the previous tasks, there is no known fairness concern that we are targeting for the CIFAR-10 task. Instead, this task allows us to examine the impact of the PW loss on overall model performance after pruning.

\begin{figure*}
  \centering
  \includegraphics[width=\textwidth]{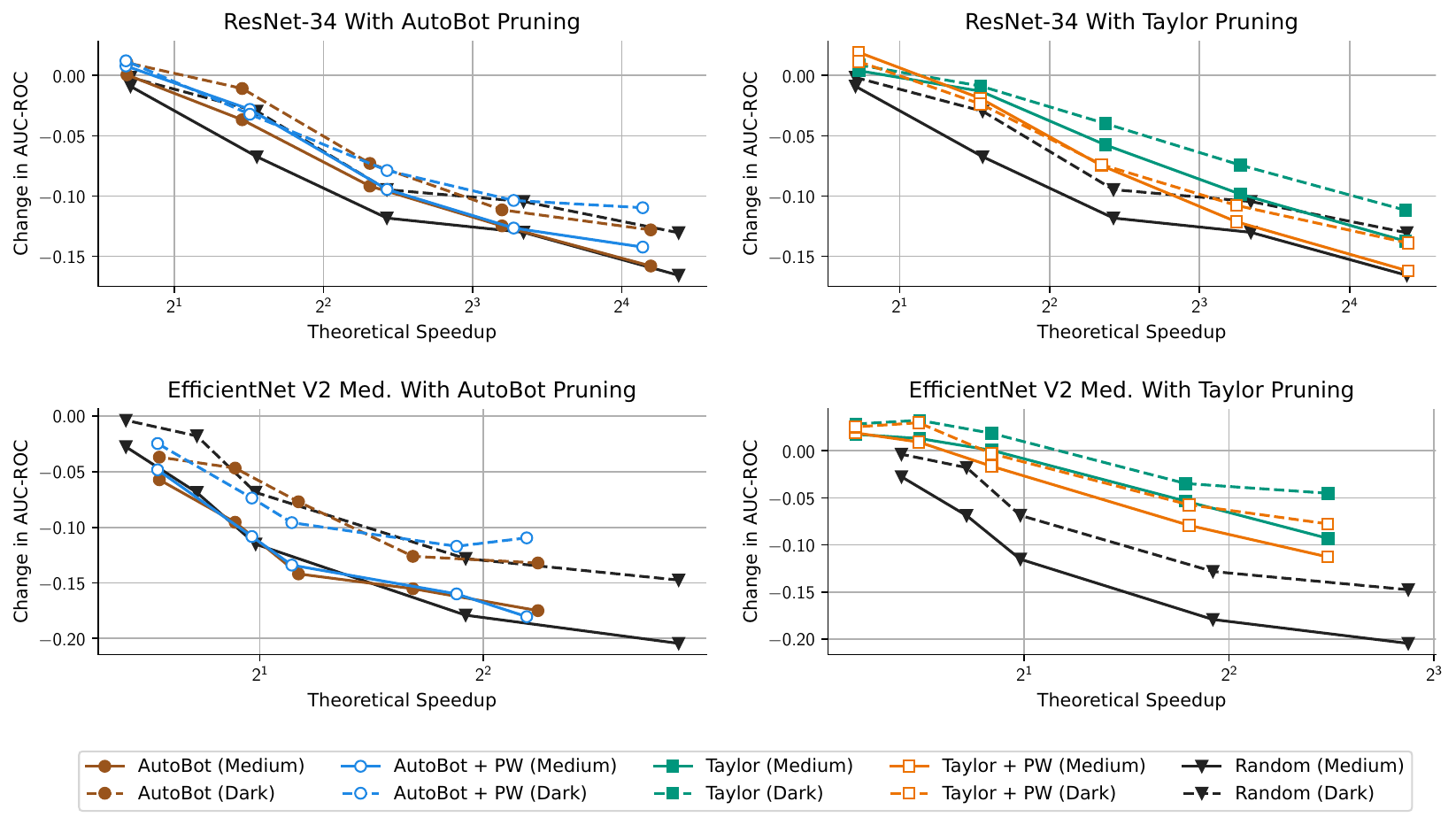}
  \vskip -0.02in
  \caption{Mean pruning performance with ResNet-34 and EfficientNet-V2 Med. models with Fitzpatrick17k dataset.}
  \label{fig:fitz17k}
\end{figure*}

\subsubsection{Pruning the CelebA Models}

We trained a ResNet-18 \cite{resnet} model and a VGG-16 \cite{simonyan2015} model for the CelebA task. The ROC-AUCs for the male and non-male subgroups of the ResNet-18 model were 0.9639 and 0.9794 respectively. The ROC-AUCs for the male and non-male subgroups of the VGG-16 model were 0.9679 and 0.9825 respectively. Both models were pruned using target theoretical speedups of 16, 32, 64, 128 and 256.

The change in ROC-AUC for all tested pruning methods for the ResNet-18 and VGG-16 models can be found in Figure \ref{fig:celeba}. Some of the VGG-16 models pruned using the AutoBot, AutoBot with performance weighting and random methods always predicted a single class. These degenerate models were excluded from the figure. All methods were able to significantly reduce the size of both models, but most of the results without performance weighting exhibited divergent performance between the male and non-male subgroups as the theoretical speedup increases.

Performance weighting was most effective for the ResNet-18 model with AutoBot pruning. We see an increase in AUC-ROC at all tested theoretical speedups for both the male and non-male subgroups. The increase for the male subgroup is substantial and the subgroup AUC-ROC scores no longer diverge as the theoretical speedup increases. At lower theoretical speedups, it was also similarly effective for the ResNet-18 model with Taylor pruning, the VGG-16 model with AutoBot pruning and the VGG-16 model with Taylor pruning. For all models and pruning methods, the PW loss appears to be more impactful for male samples than non-male samples.

\subsubsection{Pruning the Fitzpatrick17k Models}

\begin{figure*}[h!tb]
  \centering
  \includegraphics[width=\textwidth]{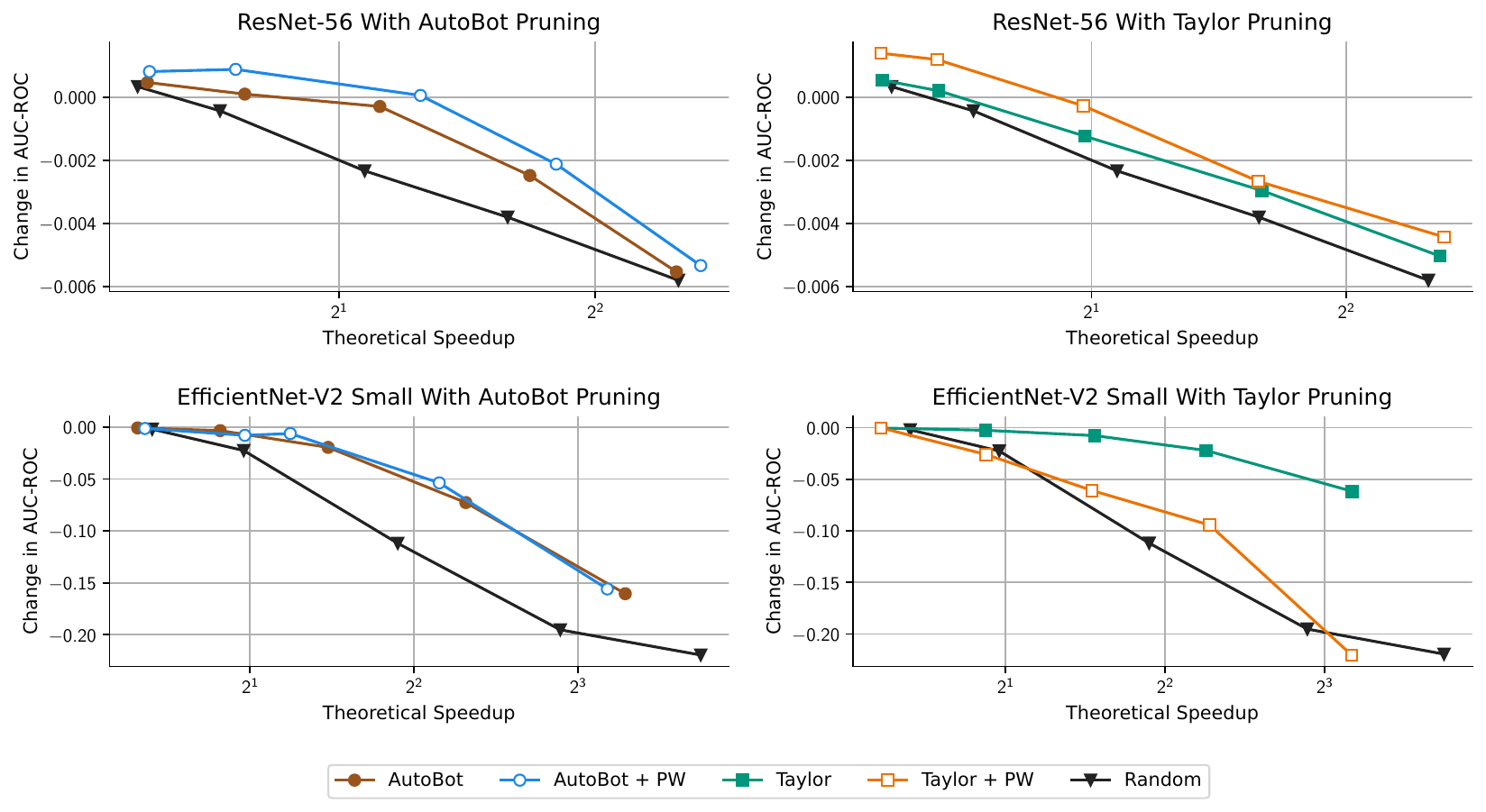}
  \vskip -0.02in
  \caption{Mean pruning performance with ResNet-34 and EfficientNet-V2 Small models with CIFAR-10 dataset.}
  \label{fig:cifar}
\end{figure*}

We trained a ResNet-34 \cite{resnet} model and a EfficientNet-V2 Medium \cite{tan2021} model for the Fitzpatrick17k task. The ROC-AUCs for the medium and dark subgroups of the ResNet-34 model were 0.8190 and 0.7329 respectively. The ROC-AUCs for the medium and dark subgroups of the EfficientNet model were 0.8516 and 0.7524 respectively. Both models were pruned using target theoretical speedups of 2, 4, 8, 16 and 32.

The change in ROC-AUC for all tested pruning methods for the Fitzpatrick17k models can be found in Figure \ref{fig:fitz17k}. Despite a bias against dark skin tones existing in the original models, we do not see divergent AUC-ROC scores as the theoretical speedup increases. The medium skin tone subgroup actually saw greater changes in AUC-ROC due to pruning. We only see slight benefits for using performance weighting with the Fitzpatrick17k models. Performance weighting increased slightly improved performance after pruning for both models with Taylor pruning at lower theoretical speedups and for the ResNet-34 model with AutoBot pruning. It had negligible or detrimental effects for all other experiments with the Fitzpatrick17k dataset.

These results indicate that performance weighting is not an appropriate solution for all datasets and models that exhibit bias. The lack of an increasing performance difference between subgroups may indicate that the pruning process was not introducing additional biases in the Fitzpatrick17k models. This is in contrast to the CelebA models for which the initial bias was small but grew due to pruning. This difference may be due to the out-of-distribution nature of the Fitzpatrick17k task. Not all fairness concerns manifest in the same manner for every dataset. Performance weighting may only mitigate biases that are introduced from the pruning process. It may not rectify biases that exist in the model before pruning.

\subsubsection{Pruning the CIFAR-10 Models}

We trained a ResNet-56 \cite{resnet} model and a EfficientNet-V2 Small \cite{tan2021} model for the CIFAR-10 task. The ResNet model was trained from scratch while the EfficientNet model was initialized using weights trained using ImageNet \cite{imagenet}. The ROC-AUCs of the ResNet and EfficientNet models were 0.9957 and 0.9995 respectively. Both models were pruned using target theoretical speedups of 1.33, 2, 4, 8 and 16.

As shown in Figure \ref{fig:cifar}, the PW loss had very different impact on the pruning processes for both models. For the ResNet-56 model, it improved performance for both tested pruning methods at all theoretical speedups. In contrast, with the EfficientNet-V2 Small model, it had a negligible impact for AutoBot pruning and a detrimental impact for Taylor pruning.

A pruning approach that prioritizes the performance of poorly classified samples can help improve the performance of models that do not have known fairness concerns. However, such an approach may not be beneficial for all models.

\begin{figure*}[h!tb]
  \includegraphics[width=\textwidth]{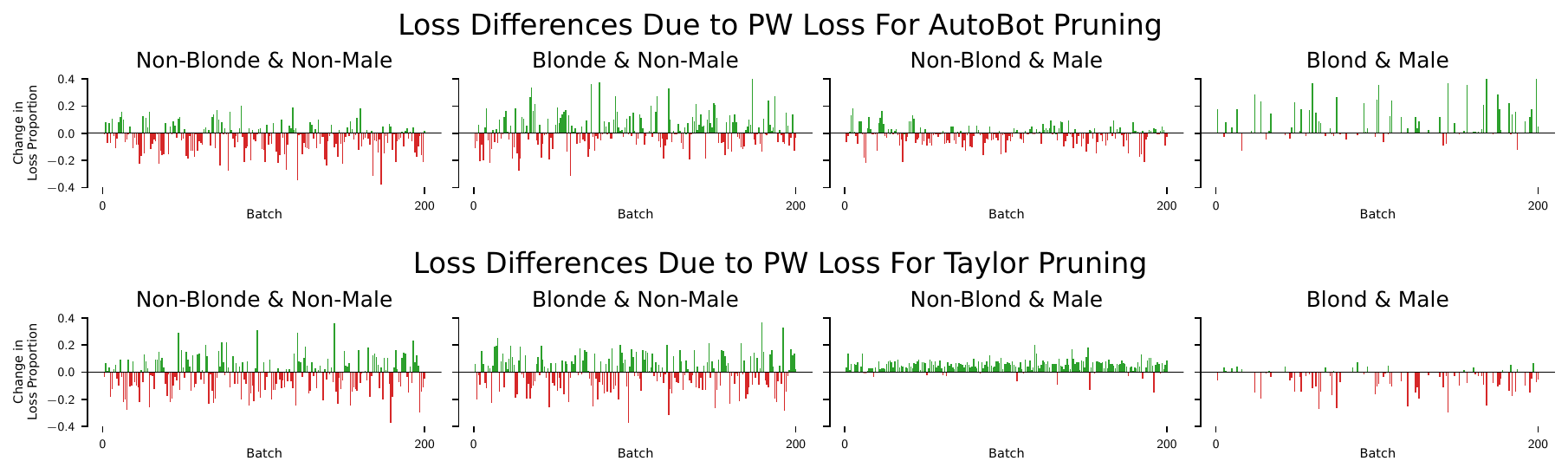}
  \centering
  \vskip -0.02in
  \caption{Change in the proportion of the batch loss due to the use of the PW loss, segmented by class and attribute.}
  \label{fig:losses}
\end{figure*}

\begin{figure*}[h!tb]
  \centering
  \includegraphics[width=\textwidth]{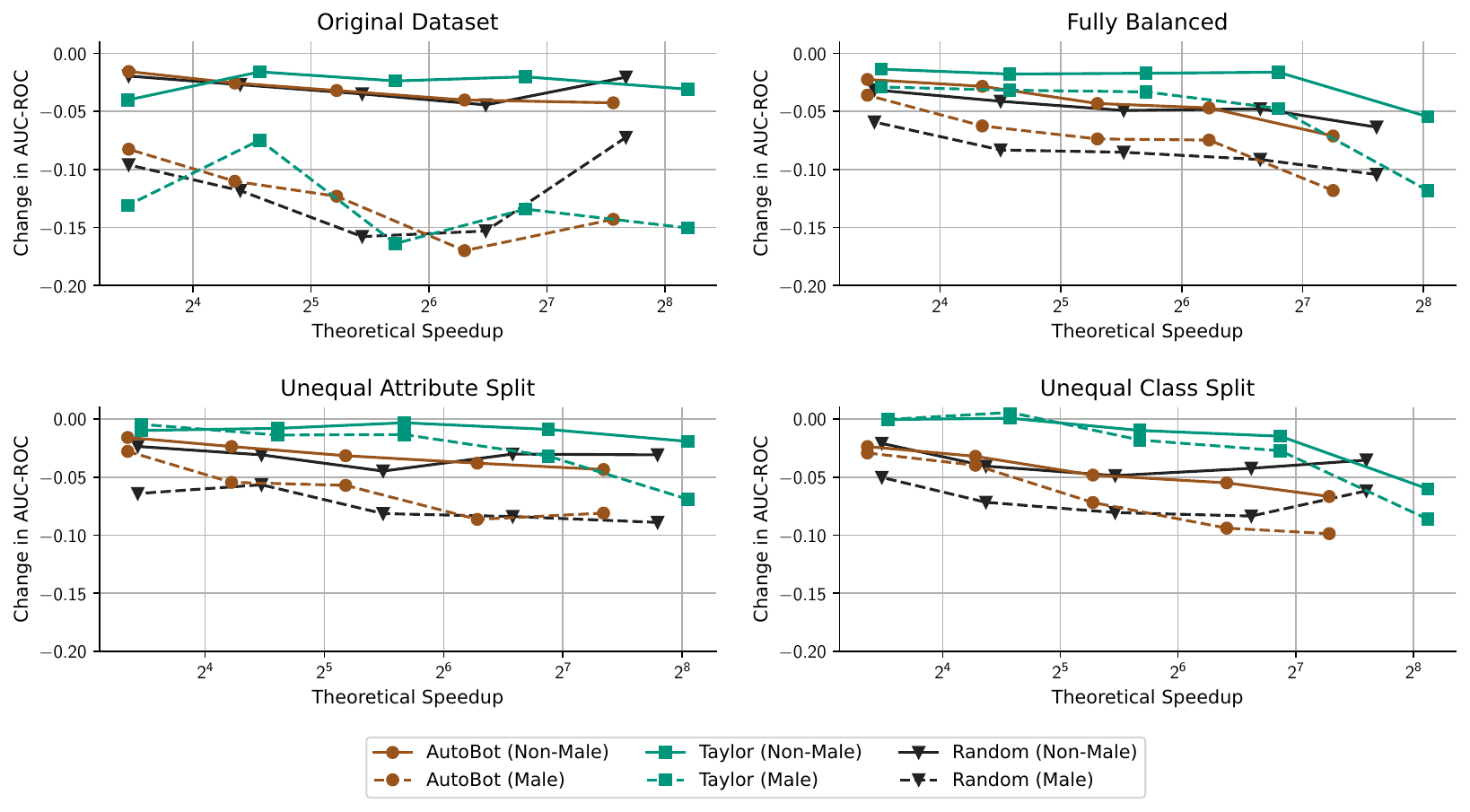}
  \vskip -0.02in
  \caption{Pruning performance with ResNet-18 models trained on subsets of CelebA dataset with alternative class and gender balances.}
  \label{fig:alt}
\end{figure*}

\subsection{Loss Analysis}

To understand the impact the PW loss has during training, we analyzed the loss values of each batch during pruning. Figure \ref{fig:losses} depicts the change in loss values for each attribute and class value for the first 200 batches of pruning the ResNet-18 CelebA model using the AutoBot and Taylor methods with the PW loss. The models were pruned using a target theoretical speedup of 16 and the same random seed was used for batch selection for all experiments.

The introduction of the PW loss to the AutoBot pruning procedure causes a clear increase in loss contribution of the blond male samples. Blond male samples are poorly represented and poorly classified by the original model. An increase in the loss contribution demonstrates that the PW loss is causing the model pruning process to place more emphasis on the performance for this group to improve fairness. In contrast, the introduction of the PW loss to the Taylor pruning procedure instead increases loss contribution of the non-blond male samples and decreases the loss contribution of the blond male samples.

Despite the difference in loss contributions, in the results depicted in Figure \ref{fig:celeba}, we did see improved fairness between male and non-male samples for both pruning methods at the utilized theoretical speedup with the use of the PW loss. This may be explained by the both methods seeing an increased loss contribution from male samples after the introduction of the PW loss, even if the specific samples which see an increase differ between pruning methods. While the PW loss did not appear to have the same effect on the two pruning methods, it was able to consistently increase the loss contribution of poorly represented samples without explicit attribute information.

\subsection{Conditions for Bias}

\subsubsection{Adjusting the Attribute and Class Balance}

From our results in Section \ref{sec:eval}, we can see that utilizing the PW loss is not necessary in all circumstances. The loss appeared to be more beneficial for models which saw increasing differences in performance between subgroups as the theoretical speedup increased.

To understand the properties of a dataset that would necessitate the use of the PW loss, we created three artificial datasets from the CelebA dataset by selected subsets of the training data. The first subset was formed using 3.41\% of the available training data such that it was fully balanced, containing an equal number of male and non-male samples as well as an equal number of blonde and non-blonde samples. The second and third subsets were formed by adding additional samples to the first subset, altering the class or gender balance. The second subset contained an equal number of blonde and non-blonde samples, but five times as many non-male samples as there were male samples. The third subset contained an equal number of male and non-male samples, but five times as many non-blonde samples as there were blonde samples. The entire test set was used to evaluate all subsets.

A ResNet-18 model was trained using each subset. The AUC-ROCs for the male subgroup are 0.9562, 0.9479 and 0.9183 for the first, second and third subsets respectively. The AUC-ROCs for the non-male subgroup are 0.9713, 0.9732 and 0.9580 for the first second and third subsets respectively. The models were pruned using the AutoBot and Taylor methods using target theoretical speedups of 8, 32 and 128. The performance after pruning for these models can be found in Figure \ref{fig:alt}.

In the results using the fully balanced subset, we do see a divergence in subgroup performance for both methods, but the divergence is smaller than what was seen when the full dataset was used.

In the results with the additional non-male samples, we see a slight increase in performance for both methods. The increase is larger at higher theoretical speedups. For the AutoBot results, the increase is greater for non-male samples than it is for male samples. Similarly, we also again see an increase in performance when we look at the subset with the additional non-blonde samples.

The difference in performance between male and non-male samples is smaller for the model with the label imbalance than it is for the model with the gender imbalance. However, a greater decrease in performance was seen for male samples for all model/method combinations, including those that were trained on data with a balanced gender split. These results indicate that the dataset composition does influence, but not fully explain, the fairness impact of model pruning.

\subsubsection{Inducing Bias in an Unbiased Dataset}

\begin{figure*}[h!tb]
  \centering
  \includegraphics[width=0.7\textwidth]{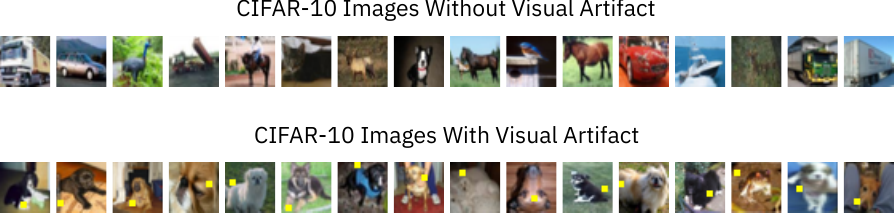}
  \vskip -2mm
  \caption{CIFAR-10 images with and without visual artifact.}
  \label{fig:cifar10_shortcut_images}
\end{figure*}

To further identify the causes of the fairness impact of pruning, we attempted to induce the effect in a dataset with no known biases. We utilized the CIFAR-10 classification task \cite{cifar10} for this purpose. 

We retrained and repruned the ResNet-56 model \cite{resnet} using two different manipulated CIFAR-10 datasets with potential sources of bias. Our first source of bias was a class underrepresentation while our second source of bias was a visual artifact that correlated with the image class. 

\subsection{Bias source: Underrepresentation}

To test if underrepresentation affects fairness during pruning we removed 80\% of the samples associated with the cat and dog classes from the training set. We then trained a new ResNet-56 model. This model had a ROC-AUC of 0.9948. The average one-vs-rest ROC-AUC of the affected classes was 0.9880. The average one-vs-rest ROC-AUC of the affected classes for the original ResNet-56 model was 0.9902.

\begin{figure*}[h]
  \centering
  \includegraphics[width=\textwidth]{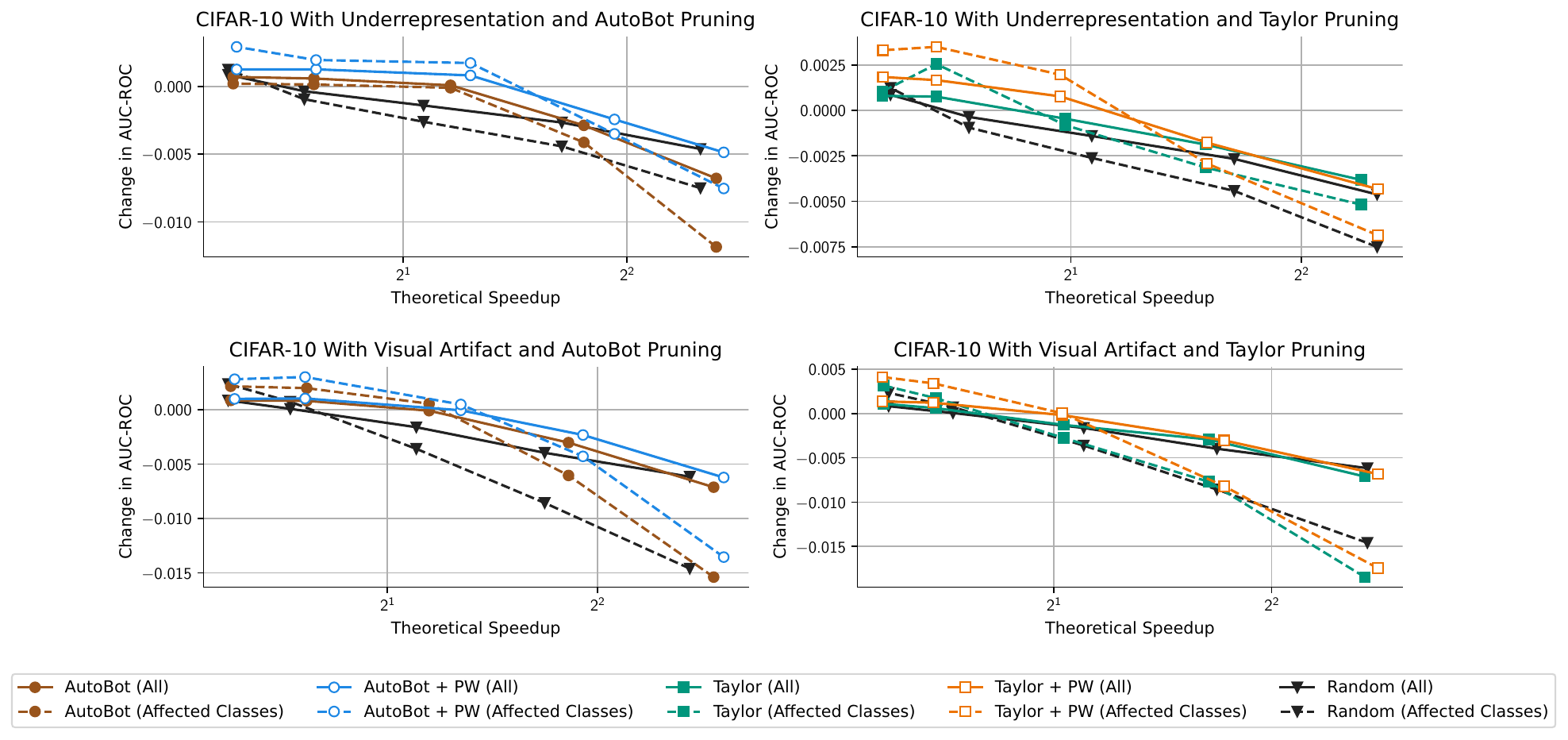}
  \vskip -2mm
  \caption[Mean pruning performance with the CIFAR-10 dataset with underrepresentation]{Mean pruning performance of the ResNet-56 model with the CIFAR-10 dataset with underrepresentation and visual artifact.}
  \label{fig:cifar10_underrep_shortcut}
  \vskip 3mm
\end{figure*}

We then pruned the new model using all three pruning methods and measured both the overall ROC-AUC and the ROC-AUC of the affected classes. The results from these experiments can be found in Figure \ref{fig:cifar10_underrep_shortcut}.

At lower theoretical speedups, the difference between the ROC-AUC of the affected classes and the overall ROC-AUC is small for all pruning methods. However, as the theoretical speedup increases, the difference grows. This effect is greater for the AutoBot method. Introducing the PW loss does appear to improve model performance and fairness at lower theoretical speedups for both pruning methods and at higher theoretical speedups for the AutoBot method.

These results indicate that underrepresentation can worsen fairness issues during pruning, but it is not guaranteed to do so. Furthermore, introducing underrepresentation into the dataset had a different effect for different pruning methods.

\subsection{Bias source: Visual artifact}

To test how the content of the image can affect fairness during pruning we introduced a visual artifact into some of the training images. Images in the dog class had a 98\% chance of having a yellow square placed in the image at a random position. Images in other classes had a 2\% chance of containing the yellow square.  This means that if a yellow square is found in an image, the image has a 84.5\% chance of being associated with the dog class. Sample images with and without the artifact can be found in Figure \ref{fig:cifar10_shortcut_images}.

We trained a new ResNet-56 model on the CIFAR-10 dataset with the visual artifact. This model had a ROC-AUC of 0.9957. The one-vs-rest ROC-AUC of the dog class was 0.9914. We then pruned the new model using all three pruning methods and measured both the overall ROC-AUC and the ROC-AUC of the dog class with and without the visual artifact. The results from these experiments can be found in Figure \ref{fig:cifar10_underrep_shortcut}.

All pruning methods saw moderate improvements in fairness due to the introduction of performance weighting. Performance weighting increased the ROC-AUC of the affected classes more than the overall ROC-AUC. This improvement was smaller at higher theoretical speedups for the Taylor method while the impact was more consistent at all theoretical speedups for the AutoBot method.

In natural images, the presence of an easy to detect feature that correlated with the classification target could have the same effect as the yellow square. This may explain the fairness issue observed with the CelebA dataset. The male property strongly correlates with the classification target of blonde hair. This correlation could have a similar effect as the correlation of the visual artifact. In both situations, performance weighting did appear to have a positive impact on fairness.

\begin{figure*}[h!tb]
  \centering
  \includegraphics[width=\textwidth]{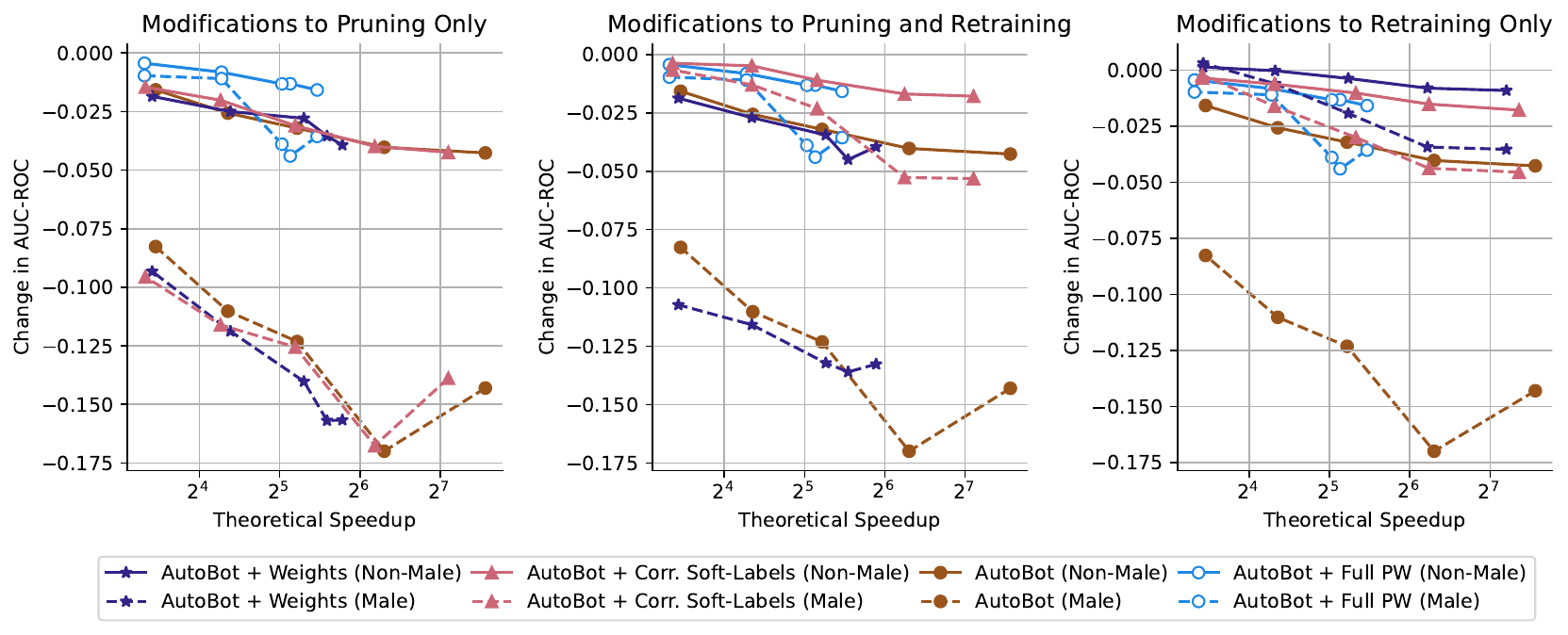}
  \vskip -0.02in
  \caption{Pruning performance with ResNet-18 models with CelebA dataset when elements of PW loss are applied independently to the pruning process (left), to the pruning process as well as the post-prune retraining process (center) and to only the post-prune retraining process (right).}
  \label{fig:ablation}
\end{figure*}

\subsection{Ablation}

To measure the effects of the components of the PW loss independently, we pruned our ResNet-18 CelebA model using the AutoBot method with only the corrected soft-labels and with only the weighting scheme described in equation \ref{eq:weights}. We applied the modifications to only the pruning process, to only the retraining process and to both the pruning and retraining processes.

The ablation results can be found in Figure \ref{fig:ablation}. Both the modifications were only effective when the retraining process was modified, indicating that simply modifying the process by which parameters are selected to be pruning is insufficient to mitigate the effects of bias. The corrected soft labels were similarly beneficial when they were applied to pruning and retraining as they were when applied to just retraining. The weighting scheme was most beneficial when applied only to the retraining process. Different parts of the performance weighted loss appear to be more impactful for different parts of the pruning process. In some situations is may be beneficial to solely apply the components of the performance weighted loss. In particular, the use of corrected soft-labels does not require the selection of any parameters, allowing it to be used in situations in which parameter selection is not feasible.

We can also see that each test either has a similar effect as the full method or little effect at all. It is possible that there is a tipping point at which any applied fairness boosting methods cause the pruning process to converge to a fair solution instead of the original unbiased solution. If the process is already converging to the fair solution, any additional fairness boosting methods would have a small effect. If the fairness boosting methods are insufficient to encourage convergence to a fair solution, the effect of the methods would be minimal.

\section{Conclusion}\label{sec:concl}

In this paper we demonstrate how model pruning can aggravate biases in convolutional neural networks and present the performance weighted loss function as a novel method for mitigating this effect. The performance weighted loss function is a simple modification that can be applied to any pruning method that uses the cross-entropy loss. Our experimental results indicate that the performance weighted loss function can help prevent model biases from becoming exaggerated in many, but not all, pruning processes. The performance weighted loss function is a useful tool for practitioners who seek to compress existing models without introducing new fairness concerns.

\section*{Acknowledgements}

This work was funded by the Natural Sciences and Engineering Research Council of Canada, the Ontario Graduate Scholarship and the Canada Research Chairs Program. We thank them for their support.

\bibliography{references}
\bibliographystyle{icml2023}

\newpage
\appendix
\onecolumn
\section{Model Training and Pruning Parameters}

To ensure transparency and enable reproducability, all parameters and procedures used to train, prune and retrain the models can be found below. All experiments were implemented using \textit{PyTorch} 1.12.1 and \textit{torchvision} 0.13.1 \cite{pytorch}. \textit{PyTorch Lightning} 1.7.1 \cite{lightning} was also used to train the models.

The ResNet-18 \cite{resnet} CelebA model was trained for 20 epochs using the AdamW \cite{adamw} optimizer with an initial learning rate of 0.0001 and a CosineAnnealingLR learning rate scheduler with $T_{max} = 20$ \cite{cosine}. A batch size of 256 was used. The model was initialized using the provided ImageNet weights from \textit{torchvision}. All parameters in layers except the final fully connected layer were frozen for the first 5 epochs after which they were unfrozen with a learning rate equal to 0.01 times the global learning rate. Early stopping was applied such that the parameters that achieved the lowest validation loss were saved after training.

The VGG-16 \cite{simonyan2015} CelebA model was trained for 10 epochs using the AdamW \cite{adamw} optimizer with an initial learning rate of 0.0005 and a CosineAnnealingLR learning rate scheduler with $T_{max} = 10$ \cite{cosine}. A batch size of 64 was used. The model was initialized using the provided ImageNet weights from \textit{torchvision}. All parameters in layers except the final fully connected layer were optimized with a learning rate equal to 0.01 times the global learning rate. Early stopping was applied such that the parameters that achieved the lowest validation loss were saved after training.

The ResNet-34 \cite{resnet} Fitzpatrick17k model was trained for 30 epochs using the AdamW \cite{adamw} optimizer with an initial learning rate of 0.001 and a CosineAnnealingLR learning rate scheduler with $T_{max} = 30$ \cite{cosine}. A batch size of 64 was used. The model was initialized using the provided ImageNet weights from \textit{torchvision}. All parameters in layers except the final fully connected layer were frozen for the first 5 epochs after which they were unfrozen with a learning rate equal to 0.001 times the global learning rate.

The EfficientNet V2 Medium \cite{tan2021} Fitzpatrick17k model was trained for 30 epochs using the AdamW \cite{adamw} optimizer with an initial learning rate of 0.001 and a CosineAnnealingLR learning rate scheduler with $T_{max} = 30$ \cite{cosine}. A batch size of 32 was used. The model was initialized using the provided ImageNet weights from \textit{torchvision}. All parameters in layers except the final fully connected layer were frozen for the first 5 epochs after which they were unfrozen with a learning rate equal to 0.01 times the global learning rate.

The ResNet-56 \cite{resnet} CIFAR-10 model was built using the implementation from \citet{idelbayev}. The model was trained for 200 epochs using the AdamW \cite{adamw} optimizer with an initial learning rate of 0.1 for the final fully connected layer and 0.01 for all other parameters. The learning rate was multiplied by a factor of 0.1 at epochs 100, 150 and 175. A batch size of 256 was used.

The EfficientNet V2 Small \cite{tan2021} CIFAR-10 model was trained for 10 epochs with an initial learning rate of 0.001 and a CosineAnnealingLR learning rate scheduler with $T_{max} = 10$ \cite{cosine}. A batch size of 128 was used. The model was initialized using the provided ImageNet weights from \textit{torchvision}. All parameters in layers except the final fully connected layer were frozen for the first 5 epochs after which they were unfrozen with a learning rate equal to 0.01 times the global learning rate.

The parameter values used for our AutoBot \cite{castells} implementation can be found in Table \ref{tab:prune_ab}. $\beta_{AB}$ refers to the parameter used by the AutoBot method to control the balance between the different terms of its loss function.

The parameter values used for our Taylor \cite{molchanov2017} implementation can be found in Table \ref{tab:prune_taylor}. $f_\textrm{prune}$ refers to the frequency of the pruning. That is, the number of batch iterations between the pruning of filters. $N_\textrm{filters}$ refers to the number of convolutional filters that are pruned in each pruning instance.

The parameter values that are used for our PW losses can be found in Table \ref{tab:prune_pw}. Other parameters were not changed when the PW loss was introduced.

After pruning, all models were retrained using the AdamW \cite{adamw} optimizer and CosineAnnealingLR learning rate scheduler with a $T_{max}$ value equal to the number of epochs. The parameter values used to retrain the models can be found in Table \ref{tab:retrain}.

\begin{table}[h]
  \centering
  \caption{Parameters used for AutoBot pruning method}
  \vskip 0.1in
  \begin{tabular}{c c c c c c c} 
    \toprule
    Dataset        & Model                & Learning Rate & Batch Size & Iters. & $\beta_{AB}$ \\
    \midrule
    CelebA         & ResNet-18             & 0.85          & 64         & 200  & 2.7  \\ 
    CelebA         & VGG-16                & 1.81          & 64         & 250  & 3.07 \\
    Fitzpatrick17k & ResNet-34             & 1.5           & 32         & 400  & 0.5  \\
    Fitzpatrick17k & EfficientNet V2 Med.  & 1.5           & 16         & 600  & 6.76 \\
    CIFAR-10       & ResNet-56             & 0.7           & 128        & 200  & 5.3  \\
    CIFAR-10       & EfficientNet V2 Small & 1.88          & 64         & 200  & 2.0  \\
    \bottomrule
  \end{tabular}
  \label{tab:prune_ab}
  \vskip -0.1in
\end{table}

\begin{table}[h]
  \centering
  \caption{Parameters used for Taylor pruning method}
  \vskip 0.1in
  \begin{tabular}{c c c c c c c} 
    \toprule
    Dataset        & Model                 & Learning Rate & Batch Size & $f_\textrm{prune}$ & $N_\textrm{filters}$ \\
    \midrule
    CelebA         & ResNet-18             & 0.01          & 64         & 5  & 1  \\
    CelebA         & VGG-16                & 0.01          & 64         & 5  & 1  \\
    Fitzpatrick17k & ResNet-34             & 0.01          & 32         & 5  & 1  \\
    Fitzpatrick17k & EfficientNet V2 Med.  & 0.01          & 16         & 4  & 8  \\
    CIFAR-10       & ResNet-56             & 0.01          & 128        & 5  & 1  \\
    CIFAR-10       & EfficientNet V2 Small & 0.01          & 64         & 5  & 8  \\
    \bottomrule
  \end{tabular}
  \label{tab:prune_taylor}
  \vskip -0.1in
\end{table}

\begin{table}[h!]
  \centering
  \caption{Parameters used for PW loss}
  \vskip 0.1in
  \begin{tabular}{c c c c c} 
    \toprule
    Dataset        & Model                 & Base Method & $\theta$ & $\gamma$ \\
    \midrule
    CelebA         & ResNet-18             & AutoBot     & 0.3      & 1   \\
    CelebA         & ResNet-18             & Taylor      & 0.8      & 0.5 \\
    CelebA         & VGG-16                & AutoBot     & 0.75     & 3   \\
    CelebA         & VGG-16                & Taylor      & 0.9      & 5   \\
    Fitzpatrick17k & ResNet-34             & AutoBot     & 0.8      & 2.5 \\
    Fitzpatrick17k & ResNet-34             & Taylor      & 0.95     & 3   \\
    Fitzpatrick17k & EfficientNet V2 Med.  & AutoBot     & 0.8      & 2   \\
    Fitzpatrick17k & EfficientNet V2 Med.  & Taylor      & 0.95     & 3   \\
    CIFAR-10       & ResNet-56             & AutoBot     & 0.5      & 1   \\
    CIFAR-10       & ResNet-56             & Taylor      & 0.5      & 1   \\
    CIFAR-10       & EfficientNet V2 Small & AutoBot     & 0.7      & 0.5 \\
    CIFAR-10       & EfficientNet V2 Small & Taylor      & 0.7      & 0.5 \\
    \bottomrule
  \end{tabular}
  \label{tab:prune_pw}
  \vskip -0.1in
\end{table}

\begin{table}[h!]
  \centering
  \caption{Parameters used to retrain models}
  \vskip 0.1in
  \begin{tabular}{c c c c c} 
    \toprule
    Dataset        & Model                 & Learning Rate & Batch Size & Duration \\
    \midrule
    CelebA         & ResNet-18             & 0.0001        & 256        & 30 epochs         \\ 
    CelebA         & VGG-16                & 0.0005        & 64         & 10 epochs         \\
    Fitzpatrick17k & ResNet-34             & 0.0001        & 64         & 30 epochs         \\
    Fitzpatrick17k & EfficientNet V2 Med.  & 0.00001       & 32         & 50 epochs         \\
    CIFAR-10       & ResNet-56             & 0.001         & 256        & 200 epochs         \\
    CIFAR-10       & EfficientNet V2 Small & 0.00001       & 128        & 10 epochs         \\
    \bottomrule
  \end{tabular}
  \label{tab:retrain}
  \vskip -0.1in
\end{table}


\end{document}